\documentclass[nonacm]{acmart}

\usepackage{blindtext}

\usepackage{xpatch}

\AtBeginDocument{
  \providecommand\BibTeX{{
    \normalfont B\kern-0.5em{\scshape i\kern-0.25em b}\kern-0.8em\TeX}}}

\begin{document}

\title{A Survey on Dialogue Management in Human-Robot Interaction}

\author{Merle M. Reimann}
\email{m.m.reimann@vu.nl}
\affiliation{
  \institution{Vrije Universiteit Amsterdam}
  \streetaddress{De Boelelaan 1111}
  \city{Amsterdam}
  \state{North Holland}
  \country{Netherlands}
  \postcode{1081 HV}
}
\orcid{0000-0003-3076-5402}

\author{Florian A. Kunneman}
\affiliation{
  \institution{Vrije Universiteit Amsterdam}
  \streetaddress{De Boelelaan 1111}
  \city{Amsterdam}
  \state{North Holland}
  \country{Netherlands}
  \postcode{1081 HV}
}
\orcid{0000-0002-1932-3200}

\author{Catharine Oertel}
\affiliation{
  \institution{Delft University of Technology}
  \streetaddress{P.O. Box 5031}
  \city{Delft}
  \state{South Holland}
  \country{Netherlands}
  \postcode{2600 GA}
}
\orcid{0000-0002-8273-0132}

\author{Koen V. Hindriks}
\affiliation{
  \institution{Vrije Universiteit Amsterdam}
  \streetaddress{De Boelelaan 1111}
  \city{Amsterdam}
  \state{North Holland}
  \country{Netherlands}
  \postcode{1081 HV}
}
\orcid{0000-0002-5707-5236}

\begin{abstract}
  As social robots see increasing deployment within the general public, improving the interaction with those robots is essential. Spoken language offers an intuitive interface for the human-robot interaction (HRI), with dialogue management (DM) being a key component in those interactive systems. Yet, to overcome current challenges and manage smooth, informative and engaging interaction a more structural approach to combining HRI and DM is needed. In this systematic review, we analyse the current use of DM in HRI and focus on the type of dialogue manager used, its capabilities, evaluation methods and the challenges specific to DM in HRI. We identify the challenges and current scientific frontier related to the DM approach, interaction domain, robot appearance, physical situatedness and multimodality.
\end{abstract}



\maketitle
\section{Introduction}\label{chap:introduction}

For humans, spoken communication is a natural way of interacting with each other, their smart speakers and even their pets. Social robots are robots that are designed specifically to interact with their human users \cite{breazeal2016social} for example by using spoken dialogue. For social robots, the interaction with humans plays a crucial role \cite{fong2002survey,baraka2020extended}, for example in the context of elderly care \cite{broekens2009assistive} or education \cite{belpaeme2018social}.
\begin{figure}[htbp]
    \centering
    \includegraphics[width=\textwidth]{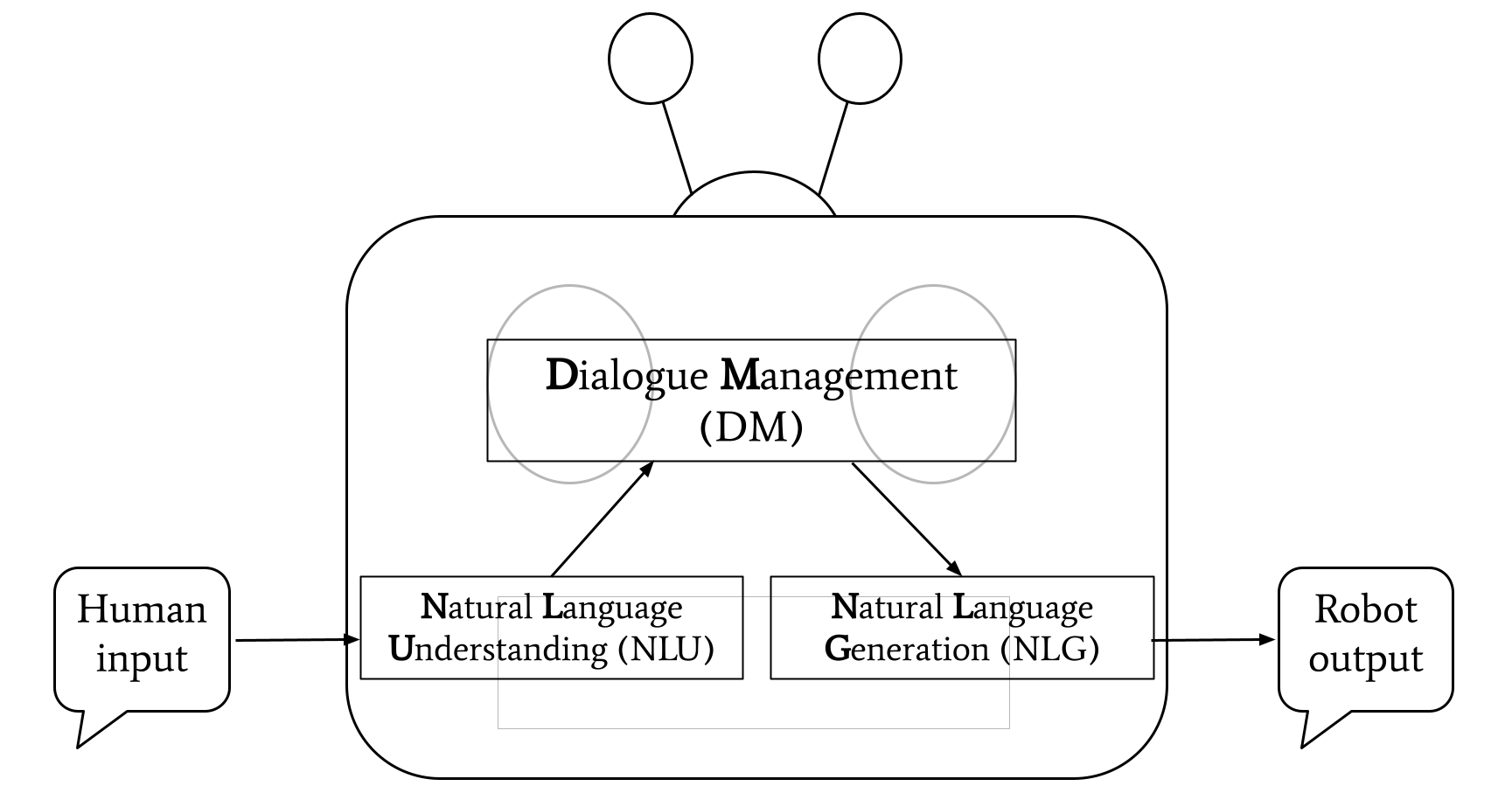}
    \caption{The integration of the dialogue manager into a spoken dialogue system.}
    \label{fig:DMinHRI}
    \Description{The agent's natural language understanding module processes the user's (spoken) input and forwards it to the dialogue manager, followed by a natural language generation module, to create the agent's output.}
\end{figure}
Robots that use speech as a main mode of interaction do not only need to understand the user’s utterances, but also need to select appropriate responses given the context. Dialogue management (DM), according to Traum and Larsson \cite{traum2003information}, is the part of a dialogue system that performs four key functions: 1) it maintains and updates the context of the dialogue, 2) it includes the context of the utterance for interpretation of input, 3) it selects the timing and content of the next utterance, and 4) it coordinates with (non-)dialogue modules. In spoken dialogue systems, the dialogue manager receives its input from a natural language understanding (NLU) module and forwards its results to a natural language generation (NLG) module, which then generates the output (see fig.~\ref{fig:DMinHRI}).

\begin{figure}[htbp]
    \centering
    \includegraphics[width=0.8\textwidth]{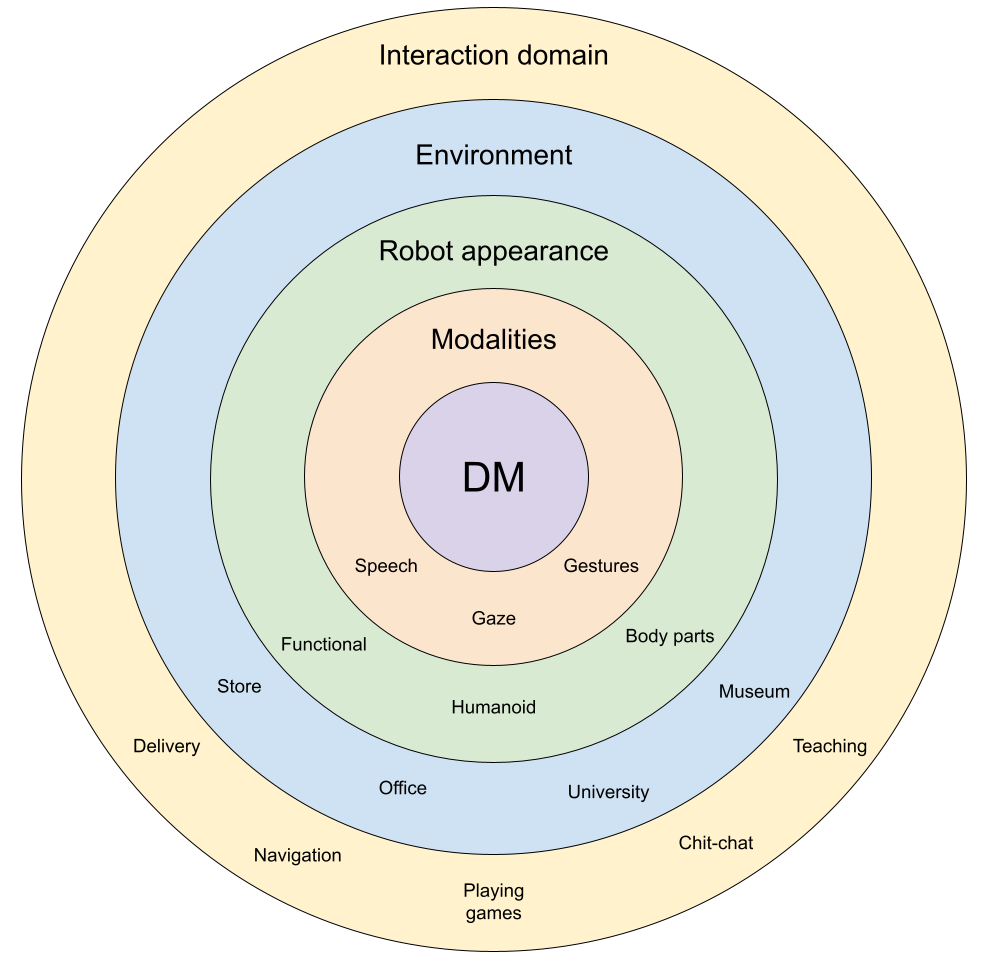}
    \caption{Dialogue management in human-robot interaction is influenced by multiple factors that show high variability. Different robot appearances can be combined with varying modalities, interaction domains and environments. Some examples are given for each factor to illustrate it.}
    \label{fig:rotating}
    \Description{The concentric circles illustrate that dialogue management is influenced by the interaction domain, environment, robot appearance and the modalities.}
\end{figure}

In contrast to general DM, DM in human-robot interaction (HRI) has to also consider and manage the complexity added by social robots (see fig.~\ref{fig:rotating}). The concentric circles of the figure describe decisions that have to be made when designing a dialogue manager for human-robot interaction. From each circle, one or more options can be chosen and combined with each other. The robot appearance, the modalities of the interaction, interaction scenarios and the physical environment influence the DM. Their combination leads to high variability, but also great complexity. While pure neural networks are used for dialogue management in non-HRI contexts \cite{brabra2021dialogue,zhao2019review}, this approach is not adopted generally for HRI, where sparse data, need for robustness and control of high-stakes interactions pose additional constraints.

While DM and HRI have not been studied extensively in combination, it is not the case for the fields individually. Harms et al. \cite{harms2018approaches} Brabra et al. \cite{brabra2021dialogue}, Deriu et al. \cite{deriu2021survey}, Zhao et al. \cite{zhao2019review} and Trung \cite{trung2006multimodal} investigate DM, but not from a robotics perspective, and Skantze \cite{skantze2021turn} looks at turn-taking, a part of DM. While there are reviews on DM in HRI, these have a specific user-group focus, such as patients suffering from dementia \cite{russo2019dialogue}. An overview of the present and future of natural language in HRI, without a specific focus on DM, is provided by \cite{liu2019review}, \cite{mavridis2015review} and \cite{marge2022spoken}. In contrast to those reviews, we focus on dialogue managers which are used in physical robots and provide a general overview of DM in HRI. With this review, we aim at giving HRI researchers an overview of the currently used dialogue management systems and their capabilities, to help them make a more informed decision when choosing a dialogue manager for their system.

\section{Review}\label{chap:methods}

We conducted a systematic review using the PRISMA protocol \cite{moher2015preferred}.
We determined inclusion and exclusion criteria, which guided the selection steps, based on our research questions:
\noindent
RQ1: Which robots are commonly used for DM? (Section~\ref{subsec:Robots})\\
RQ2: Which types of dialogue managers are commonly used in HRI, and what are the reasons? (Section~\ref{subsec:DMs})\\
RQ3: Which capabilities do current dialogue managers used in HRI have? (Section~\ref{subsec:Capabilities})\\
RQ4: How are dialogue managers in HRI evaluated? (Section~\ref{subsec:Evaluation})\\
RQ5: What are the main challenges for DM in HRI? (Section~\ref{chap:discussion})

We used Scopus, IEEE and ACM for the literature search using the search terms ``Robot AND 'mixed initiative'" and ``Robot AND ('dialog manager' OR 'dialogue manager')" resulting in 949 papers (ACM: $556$, Scopus: $278$, IEEE: $115$). Performing a second search with the more general search terms ``Robot AND (dialog OR dialogue)", but a restriction to relevant conferences (HRI, IUI, ICMI, AAMAS, SIGGRAPH, HAI, SIGDIAL, RO-MAN, IROS, INTERSPEECH, DIS, MHAI, AAAI) led to another 504 papers (ACM: $62$, Scopus: $350$, IEEE: $92$). To make sure that also papers that use the term 'interaction' instead of 'dialogue' in abstract and title are included, we performed a third search with the search term ``'social interaction' AND robot AND speech". This led to 209 (ACM: $20$, Scopus: $134$, IEEE: $55$) papers.
All papers were imported into Rayyan \cite{Ouzzani2016} for further processing. 

The formal exclusion criteria were chosen to filter out papers shorter than $4$ pages, reviews, demonstrator papers, results which are not scientific papers or did not include a physical robot. Furthermore, we excluded papers written before $2005$ since the NLU modules and DM capabilities have improved a lot since then. Additionally, the papers needed a focus on spoken DM or DM capabilities of the dialogue manager. DM capabilities are the conversational abilities the dialogue manager possesses. We decided to scope the review to only include papers that describe a whole dialogue manager. While other components like dialogue state trackers do help the dialogue manager, by providing information that the dialogue manager can then use to decide what its next action should be, they are out of scope for this survey. 
After duplicate deletion and keyword-based filtering (``dialog", ``dialogue", ``dialogs", ``dialogues", ``conversation", ``conversational", ``conversations" and ``discourse"), all authors independently performed a manual abstract screening of the remaining $753$ papers, excluding papers that do not sufficiently focus on spoken human-robot interaction. Conflicting decisions were discussed until an agreement was reached. A subsequent paper screening led to a total of $68$ papers analysed for this review.

\subsection{Robot appearance}\label{subsec:Robots}
\begin{figure}[htpb]
\centering
\includegraphics[width=0.8\textwidth]{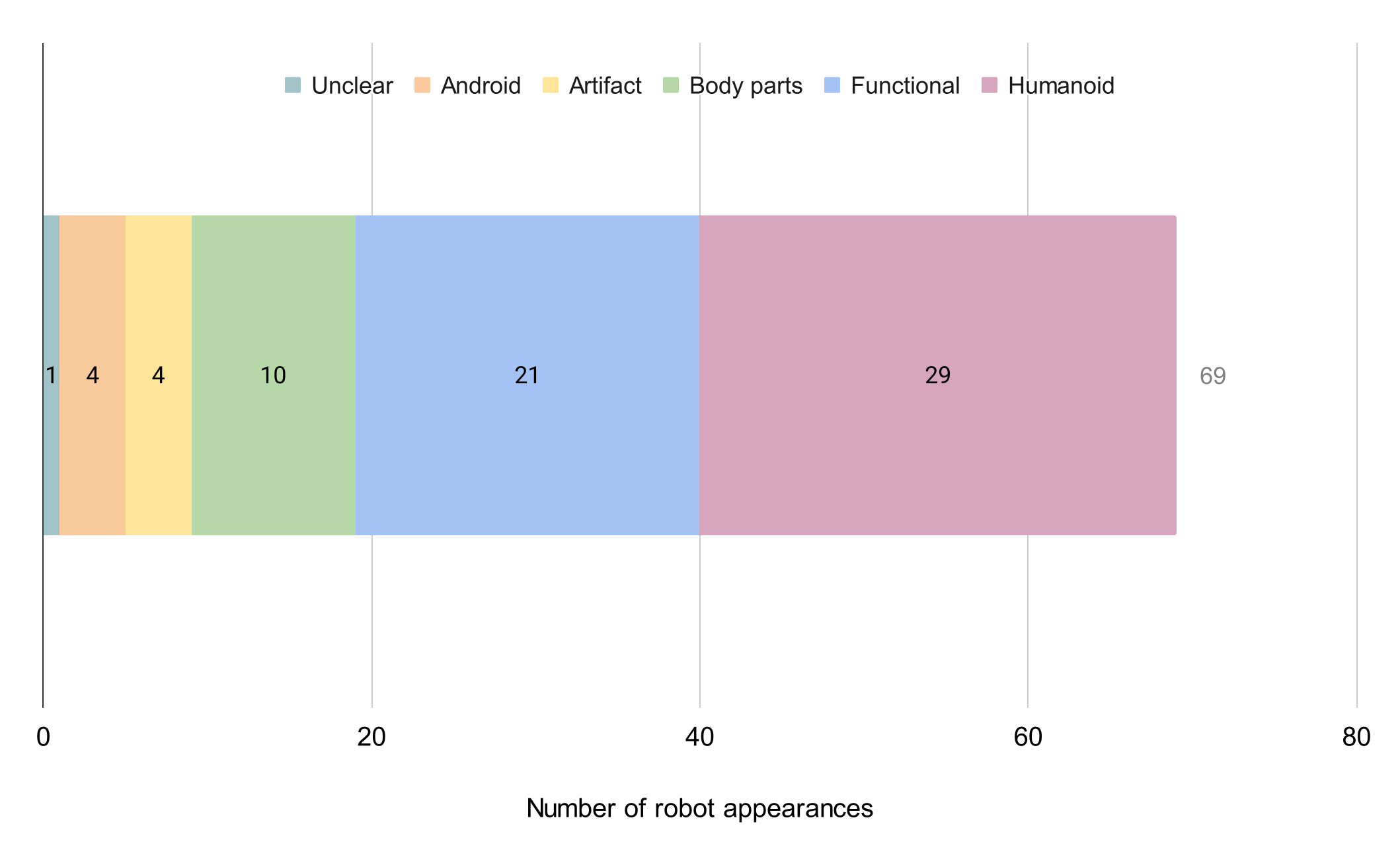}
\caption{Appearance of the robots used in the surveyed papers.}
\Description{The plot shows the distribution of appearances in the 68 surveyed papers. 29 robots are humanoid, 21 have a functional appearance, 10 are body parts, 4 are artifact shaped, 4 are android robots and one was not described in sufficient detail to classify it.}
\label{fig:robots}
\end{figure}

We classify the $69$ robots found in the $68$ papers by their appearance using the taxonomy in \cite{baraka2020extended}. Figure~\ref{fig:robots} shows the resulting distribution of types of robots found. \textit{Humanoid robots} ($41\%$) are one of the more dominant categories in spoken human-robot interaction. Within this category the NAO robot, used in $10$ papers, is the most commonly used humanoid robot \cite{chai2014collaborative,cuayahuitl2014nonstrict,lison2015hybrid,lison2012probabilistic,bohus2014managing,lison2013model,gervits2018pardon,devillers_multimodal_2015,ondas_multimodal_2017,aly_model_2013}. The Pepper robot \cite{ion2020dialog,zeng2018eliciting, nuccio_interaction_2018} and Maggie \cite{alonso-martin_multimodal_2013, alonso2015augmented,gorostiza2010natural} are used in three papers each. A robot bartender \cite{petrick2012social, foster2012two}, Armar 3 \cite{prommer_rapid_2006,holzapfel_dialogue_2008} and the PR2 robot \cite{gervits2020s,williams2015going} are mentioned twice while all other humanoid robot appearances occur just once  \cite{dino2019delivering,torrey2006effects,lee2010hybrid,jiang2011configurable,matsuyama2010framework,chao_timed_2016,li_computational_2006}.
The second biggest category consists of \textit{functional robot} appearances ($30\%$), the design of those robots is influenced by the task they are used for. The three Segway-based robots in this category are used for delivery and navigation \cite{amiri2019augmenting,thomason2015learning,zhang2015corpp}. Mobility plays a role for all robots in this category as they are used to help with information and action requests based on instructions \cite{quindere2007dialogue,quindere2007information,quindere2013evaluation,edirisinghe2018application,chen2010developing,rybski2007interactive,lee2008implementation,lucignano_dialogue_2013,aguilar_integrating_2009,aviles2010integrating}, act as a guide \cite{vogiatzis2008framework,li2006dialog,rosenthal2010mixed,gervits2021classification,li_computational_2006, lu_leveraging_2017} or learn location names \cite{nakano2010grounding,funakoshi2007robust}.

For the robots based on body parts, two robotic heads were used: Flobi \cite{carlmeyer2014towards,kipp2014dynamic} and Furhat \cite{campos2018challenges,skantze2013exploring,johansson2015opportunities}. In contrast to robotic heads, that do not have manipulators, robotic arms are used for object manipulation and grasping. We found robotic arms in five of the papers \cite{peltason2010pamini,peltason2012structuring,shen2016speech,she2014back,sugiura2010active}. Robots based on other body parts were not found in any of the papers.
All of the artifact shaped robots found in the papers are robotic wheelchairs \cite{doshi2007efficient,hemachandra2014information,hemachandra2015information,vale2013tacit}, which are used for navigation. The papers using android robots, all use ERICA as the robot \cite{lala2017attentive,lala2018evaluation,milhorat2019conversational,inoue2020attentive}.
One of the robots was not described sufficiently to classify it \cite{gieselmann2007problem}.

Based on the diversity of robot types used, we can conclude that the use of spoken dialogue systems is not restricted to specific robots with certain shapes or features, which adds to the high variability of factors influencing the DM (see fig.~\ref{fig:robots}).

\subsection{Types of DM in HRI}
\label{subsec:DMs}
\begin{figure}
    \centering
    \includegraphics[width=1\textwidth]{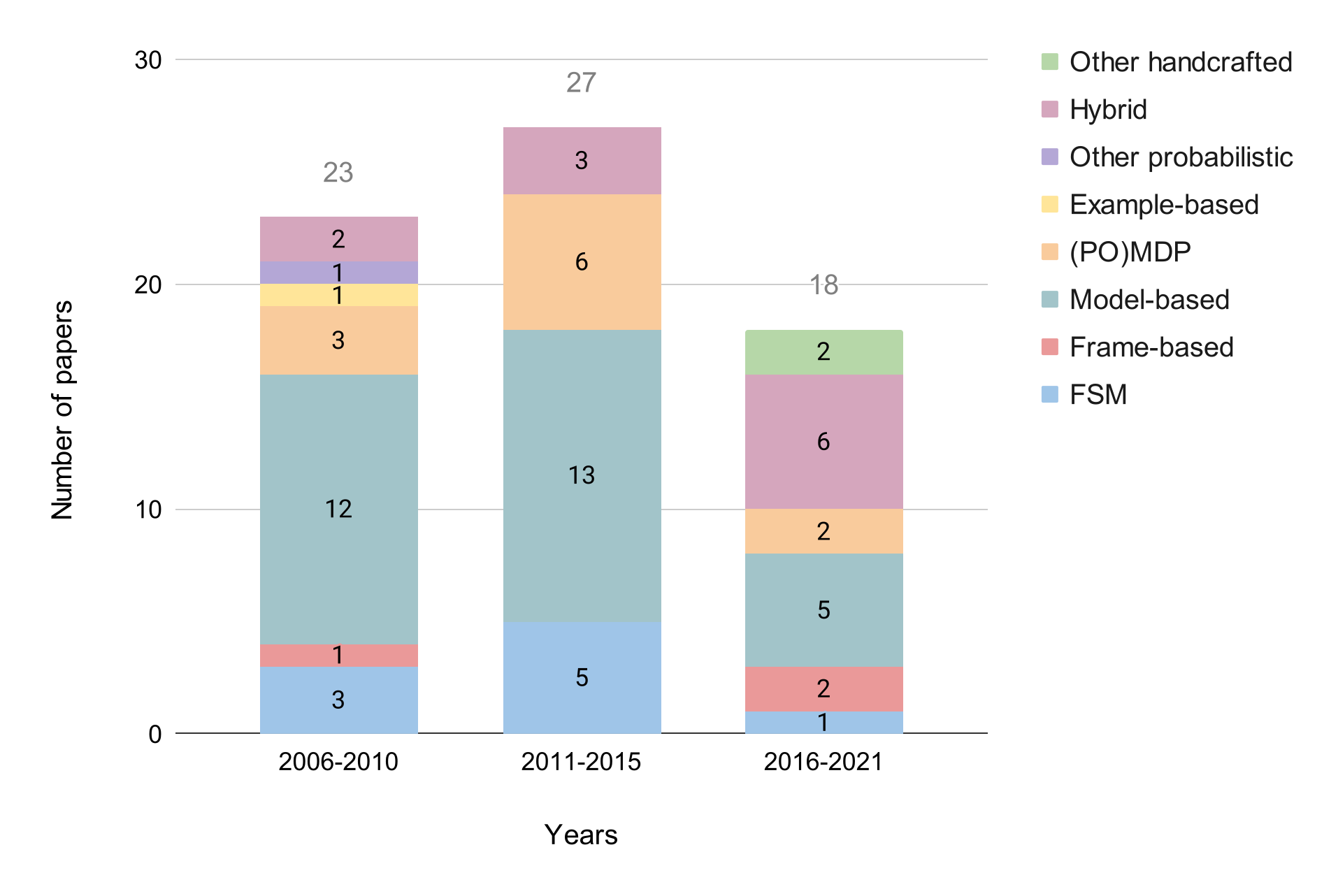}
    \caption{The dialogue management approaches over the years used in the surveyed papers.}
    \label{fig:DM}
    \Description{From 2006-2010 23 papers were selected, from 2011-2015 27 and from 2016-2021 18. In the first two intervals model-based DM was the most common one, in the third interval hybrid methods had the greatest popularity.}
\end{figure}

While the first dialogue management approaches starting in the 60s were purely handcrafted, there has been a development into the direction of probabilistic and hybrid approaches. We use the framework provided by Harms et al. \cite{harms2018approaches} as a basis for the classification of dialogue managers into handcrafted, probabilistic and hybrid approaches. In total, $45$ of the $68$ papers use handcrafted approaches, $12$ use probabilistic ones and $11$ make use of a hybrid approach (see fig.~\ref{fig:DM}). From 2006 to 2010, $17$ of the $23$ included dialogue managers were handcrafted, with model-based approaches making up over $50\%$ in total. Model-based approaches stayed popular from 2011 to 2015, but (partially observable) Markov decision processes ((PO)MDPs) saw an increased use as well. From 2016 onward, those two approaches were surpassed by hybrid ones that can combine the advantages of both.

\subsubsection{Handcrafted approaches}
The main difference between the different handcrafted approaches lies in the information encoded in the states of the dialogue manager. Finite-state machines (FSMs) are a simple approach to handcrafted DM, where the states of the FSM directly correspond to the dialogue states. A simple FSM approach is used in \cite{rybski2007interactive}, where the focus of the interaction lies on the learning of tasks through dialogue. \cite{lee2008implementation} use a composition of multiple FSMs to manage the different interactions their robot is offering. However, nowadays they are mostly used as building blocks in combination with other additional functionalities in the surveyed papers. For example, \cite{edirisinghe2018application} uses an FSM with predefined dialogues for knowledge acquisition, but aims at adding an autobiographical memory to the robot. Dialogue managers based on state charts, which are used to implement interaction patterns, are used in \cite{carlmeyer2014towards,kipp2014dynamic,peltason2010pamini,peltason2012structuring}. Interaction patterns are based on general sequences that can be found during dialogues and are defined before the interaction. Another dialogue manager based on state charts is IrisTK, which adds recursion \cite{skantze2013exploring,johansson2015opportunities}. Building upon IrisTK, \cite{johansson2015opportunities} add a data-driven turn-taking approach to the system.

To allow the human user to take more initiative and provide information more flexibly in dialogues, frame-based systems have been used \cite{bobrow1977gus}. Frame-based approaches employ empty slots, typically information units required to fulfill the users request, that can be filled at any point of the conversation. \cite{gorostiza2010natural} present such a system, where multiple slots can be filled at a time, for action requests for the robot Maggie. A frame-based approach is also adopted in \cite{zeng2018eliciting}, where the user’s food preferences are determined in an interview-like conversation by using slot-filling.

The third handcrafted approach is the model-based approach. In addition to the frames with slots, a model-based dialogue manager has more complex states. Those states include some form of model, for example about the user, the environment, the situation or the context. Model-based dialogue managers often consider the user model, including the user’s goals and intention \cite{chai2014collaborative,williams2015going,thomason2015learning,quindere2007dialogue,quindere2013evaluation,vogiatzis2008framework,rosenthal2010mixed}. User models that influence the dialogue management can also be based on displayed emotion \cite{alonso-martin_multimodal_2013, devillers_multimodal_2015} or personality \cite{aly_model_2013} of the user. For long-term interaction, including a model of the conversation history is especially relevant and enables the dialogue manager to learn from past interactions. Cobot’s dialogue manager described in \cite{rosenthal2010mixed} is designed for a long-term interaction, and therefore uses the conversation history to avoid repetitive dialogue or unnecessary offers. In \cite{dino2019delivering}, the robot collects and stores user data that influences later conversations with the user. \cite{li2006dialog, li_computational_2006} use a stack to keep track of unfinished conversation sequences for tracking the context during the interaction. The dialogue manager used in \cite{nuccio_interaction_2018} groups utterances into contexts, which then can be focused during different points of the conversation.

The Information State Update approach \cite{traum2003information} used by \cite{quindere2007dialogue,quindere2007information,quindere2013evaluation}, stores and uses amongst other things the dialogue context, information about the possible referents of pronouns, and information about the current turn and initiative. \cite{she2014back, aguilar_integrating_2009,aviles2010integrating} also use the dialogue context and user’s intention, in addition to objects referenced in the environment, to determine the next dialogue move. Multiple expert modules, activated based on domain scores, are used by \cite{nakano2010grounding} and \cite{funakoshi2007robust}.

Plan-based systems, which belong to the model-based approaches, treat the conversation as a planning problem and aim at reaching a goal state by selecting from predefined actions. Plan-based dialogue management is used by \cite{petrick2012social}, \cite{alonso2015augmented} and \cite{foster2012two}, who use the planner for the dialogue and action generation. Other plan-based systems are used by \cite{vale2013tacit} for navigation of a robotic wheelchair and by \cite{gervits2018pardon}, with additional rule-based speech overlap management. \cite{chen2010developing} uses answer set programming for their planning module, which can generate a high-level plan based on the user’s request and a low-level plan for the motion planner. In \cite{campos2018challenges}, a goal, based on novelty, dialogue history, relevance and diversity, is used for the utterance selection, while \cite{gervits2021classification} performs logical inference over the goals and beliefs to this end.

The dialogue manager used by \cite{jiang2011configurable} can be used as a model-based one, but a simplified version can also be used as a frame-based or FSM dialogue manager. Two dialogue managers do not fit the classification and were therefore classified as \textit{other handcrafted}. \cite{chao_timed_2016} used timed petri nets to improve mutltimodal interaction, especially in respect to turn-taking. Petri nets are state-transition networks that have tokens, which determine the current configuration state and whether a transition is enabled. In addition to utilizing this approach, timed Petri nets use real time transitions of the tokens to be able to model the configuration according to the input. \cite{ondas_multimodal_2017} use a dialogue manager that uses XML files for generating the dialogue in Slovac languages.

\subsubsection{Probabilistic approaches}
In contrast to handcrafted approaches, probabilistic approaches learn the dialogue policy directly from data. A probabilistic approach used in only one surveyed paper is the example-based approach \cite{torrey2006effects} that looks up the most similar utterance in its database, consisting of numerous conversations, and uses the answer given there as its own.

A popular probabilistic approach used in DM are (partially observable) Markov decision processes ((PO)MDPs). MDPs model the dialogue as a Markov process, moving from one state to another based on learned transition probabilities. For POMDPs, the current state is not known and is estimated based on observations (e.g. perceived actions and utterances) and their probabilities. The robotic wheelchairs used in \cite{hemachandra2014information} and \cite{hemachandra2015information} use dialogue to check their spatial semantic representation and make sure that it corresponds to the real environment, by asking questions about the environment. The POMDP dialogue manager in \cite{amiri2019augmenting} rewards successful conversations, leading to task completion, while additional questions, confirmations and unsuccessful conversations are penalised. Two POMDPs, one for the DM and another one for deciding when to update the knowledge base, are used for improving the augmentation of the knowledge on an as-needed basis. \cite{cuayahuitl2014nonstrict} allow flexible navigation between the different sub-dialogues, to enable the user to change their mind about the order in which they want to do certain tasks. A Bayesian approach to learning the best dialogue strategy is used by \cite{doshi2007efficient} for a robotic wheelchair. Integrating reasoning with probabilistic common sense knowledge enables the robot in \cite{zhang2015corpp} and \cite{lu_leveraging_2017} to ignore irrelevant variables while planning. Another approach for limiting the state space is used in \cite{lison2012probabilistic}, where probabilistic rules are used for Bayesian networks.
(PO)MDPs are also used for multimodal dialogue, where they learn the strategies from multimodal instead of pure dialogue data \cite{prommer_rapid_2006,holzapfel_dialogue_2008,lucignano_dialogue_2013}.

We did not find a system purely based on an end-to-end learning approach, but end-to-end approaches are in fact used as parts of hybrid systems (see section~\ref{subsec:DMs}, Hybrid approaches). A probabilistic approach which does not fit into either category described earlier is used by \cite{sugiura2010active}. They generate all possible linguistic expressions for an action, which is possible since they use a simplified sentence structure. Then they use expected log loss reduction and Bayesian logistic regression, predicting which utterance should be selected. Using the best score, an utterance is selected and produced. After the production, the appropriateness of the utterance is rated based on the user’s reaction and the example is added to the training set. This leads to active learning during the conversation.

\subsubsection{Hybrid approaches}
Instead of focusing on either a handcrafted or data-driven approach, in recent years it has become more popular to use a combination of the two. These so-called hybrid approaches can be fitted according to the needs of the target application and the availability of data. The option to combine approaches can compensate for their weaknesses and utilise their strengths. Hybrid approaches are often used when the dialogue manager is distributed in a variety of submodules which focus on different aspects of the dialogue management, like turn-taking, engagement management or problem handling. \cite{lala2018evaluation} combine a finite-state turn-taking machine with a deep neural network, in order to improve the dialogue manager’s turn-taking capabilities. Turn-taking is also integrated into the otherwise model-based dialogue manager in \cite{gervits2020s} by adding a combination of incremental processing and lexical prediction. In \cite{bohus2014managing} an FSM approach is combined with self-supervised learning to handle disengagement. To handle problems occurring during the dialogue more efficiently, \cite{gieselmann2007problem} add an FSM with four states (Start, Normal, Help, Error) to the dialogue manager ARIADNE \cite{holzapfel2005towards}. \cite{shen2016speech} also use an FSM for tracking the general state of the dialogue system, but make use of external services for questions and declarative utterances.

\cite{lala2017attentive} use logistic regression to select so-called experts, which can have different strategies, while \cite{inoue2020attentive} rely on a priority system in combination with additional triggers for a backchannel and backup question module. Backchannels are short responses that are integrated into the conversation while listening, to show engagement (e.g. ``hmm", ``yeah"). In \cite{milhorat2019conversational} a dialogue act tagger based on supervised machine learning (Support Vector Machines) is used to decide whether a statement response should be generated, based on either a decision tree or a response using an example-based approach. If neither have a sufficiently high confidence score, a backchannel is produced instead.

It is also possible to make use of probabilistic rules for DM  to reduce the parameter space by including expert knowledge \cite{lison2015hybrid}. A combination of a POMDP and a model-based approach is used in \cite{lison2013model}, where the user’s goal and actions are modelled using Bayesian inference. A combination of an example-based and agenda-based approach is used in \cite{lee2010hybrid}. While the agenda graph is designed manually, the examples which are mapped to the nodes are learned from human-human dialogues.

\subsection{Communication capabilities in HRI}
\label{subsec:Capabilities}
We list common capabilities that were explicitly mentioned in the reviewed papers in table 1. \paragraph{Confirm} Explicit confirmations by the robot for establishing common ground are important, especially for task-based interaction, to increase task success \cite{thomason2015learning}. Common ground describes shared beliefs, for example about the environment or the state of the conversation \cite{stalnaker2002common}, which is relevant, if the human and robot have to achieve a task together. Example use-case scenarios include movement instructions \cite{gorostiza2010natural,gervits2021classification}, receptionist tasks \cite{jiang2011configurable,quindere2013evaluation,holzapfel_dialogue_2008}, grasping \cite{peltason2012structuring}, delivering items \cite{thomason2015learning} or the establishment of common ground in situated interaction \cite{chai2014collaborative}.

\begin{table}[t]
    \centering
    
    \caption{Common conversational capabilities observed in the 68 surveyed papers.}
    \begin{tabular}{l c}
        \toprule
        Conversational Capability & Amount\\
        \midrule
        Confirm & 13 \\
        Ask for confirmation & 24 \\
        Ask for repetition/rephrasing & 10 \\
        Ask clarification questions & 17 \\
        Reference resolution/Spatial grounding & 24 \\
        Acquire knowledge & 21 \\
        \bottomrule
    \end{tabular}
    \label{tab:capabilities}
\end{table}

\paragraph{Ask for confirmation} A clear trend across more than $\frac{1}{3}$rd of all papers is the robot’s ability to ask for confirmation. This can happen either explicitly (e.g. ``Do you want...?", with an option for the user to (dis-)confirm) or implicitly (e.g. ``I will now...", with an option to cancel the action if it is not the desired one). \cite{peltason2010pamini,peltason2012structuring} use a correctable information request pattern for implicit confirmations, and an explicit confirmation after an Information Request pattern, for integrating confirmation questions into their task-based interaction. Confirmations by the human can be used to ensure common ground about the environment \cite{chai2014collaborative, skantze2013exploring,holzapfel_dialogue_2008}, to elicit information about user preferences \cite{zeng2018eliciting,lu_leveraging_2017}, to confirm the correctness of past actions \cite{lison2012probabilistic} or to make sure that the conversation is not terminated too early \cite{rybski2007interactive}. At the same time, unnecessary confirmation or clarification questions can make the conversation inefficient and increase user frustration and are therefore often associated with a cost in reinforcement learning settings \cite{amiri2019augmenting,lison2012probabilistic,hemachandra2015information,zhang2015corpp,doshi2007efficient}. 
For example, robots which navigate or deliver items \cite{lison2015hybrid,amiri2019augmenting,zhang2015corpp,doshi2007efficient}, or make appointments and book something \cite{jiang2011configurable,quindere2013evaluation} often ask for confirmation, since a later correction after performing the action is more costly. To avoid unnecessary confirmation questions, \cite{lison2013model} assigns a negative value to the reward function of their reinforcement learning dialogue strategy for additional confirmations, while the robot in \cite{sugiura2010active} learns when and how to confirm through active learning. Since robots have to deal with challenging environments with noise, confirmation questions are used, when the confidence that the utterance was understood correctly is low \cite{williams2015going,gorostiza2010natural,quindere2007information,quindere2013evaluation,nakano2010grounding}.

\paragraph{Ask for repetition/rephrasing} In case of speech recognition problems, a repetition of the utterance can already help resolving them, while rephrasing gives the human the option to produce a different utterance which is easier to understand for the robot than the first one. To determine whether repetition is needed, different strategies are available. Robots ask users to repeat an utterance when the recognition is below a threshold \cite{gorostiza2010natural}, no match for the input was found \cite{gervits2021classification,carlmeyer2014towards} or in case of underspecification \cite{lison2015hybrid,lison2013model}. If no grounding is achieved for object names that are referred to by a gesture accompanying a pronoun, the robot in \cite{li2006dialog} asks for repetition. Like repetition, rephrasing can be used when the dialogue manager finds no match for the input \cite{ion2020dialog,lee2010hybrid,gervits2021classification}, if the dialogue manager is uncertain either about the dialogue state \cite{amiri2019augmenting} or about the correct recognition of the utterance \cite{gorostiza2010natural}.

\paragraph{Ask clarification questions} Clarification questions are asked in those cases where the robot is unsure if it has understood the utterance correctly \cite{amiri2019augmenting,thomason2015learning,quindere2007dialogue,quindere2007information,chao_timed_2016} or when it is missing information \cite{zeng2018eliciting,gorostiza2010natural,gervits2021classification,gieselmann2007problem,holzapfel_dialogue_2008}. Additional information to make complete plans \cite{chen2010developing} or determine the current context \cite{williams2015going} can be acquired through clarification questions. If most parts of a user request are already specified and only some need additional clarification, \cite{carlmeyer2014towards} suggest to let the robot already start the action while clarifying the missing parts.

\paragraph{Reference resolution/spatial grounding} Something which distinguishes robots from conversational agents is that they are physically embodied and situated in an environment. That is why situational grounding is important for robots that, for example, engage in navigation \cite{lison2015hybrid,rosenthal2010mixed,hemachandra2014information,hemachandra2015information}, deliveries \cite{thomason2015learning} or give a tour in the environment \cite{rybski2007interactive,li2006dialog,peltason2010pamini}. Situational grounding refers to the robot's ability to ground concepts and referenced objects in the real world ("Give me the red cup" would refer to a specific red cup that is present in the environment, not to the concept of a red cup in general).
The robot in \cite{chai2014collaborative} engages in a naming game of objects in the environment with the user and focuses on establishing common ground. To do so, it makes its knowledge explicit, such that the human can correct it if necessary. Robotic arms that grasp and move objects, need to identify the object the user is talking about correctly \cite{peltason2012structuring,she2014back,sugiura2010active}. For multimodal dialogue where the human refers to objects in the environment that are related to the robot's task, the robot also needs to be aware which object the human is referencing using the available modalities (e.g. gestures, speech, gaze) \cite{holzapfel_dialogue_2008,prommer_rapid_2006,aguilar_integrating_2009,aviles2010integrating}.

\paragraph{Acquire knowledge} If the human is talking to the robot about previously unseen or unheard objects or people, the robot should be able to make sense of them. Knowledge augmentation allows for new knowledge to be added to the knowledge base of the robot.  For navigating robots, questions about their environment improve their knowledge about the spatial semantic representation \cite{hemachandra2014information,hemachandra2015information}. In the delivery domain, knowledge augmentation is used to learn names of objects or people which have to be delivered/should receive the delivery but are not yet in the knowledge base \cite{amiri2019augmenting,thomason2015learning}. Another context in which knowledge from previous interactions and users is used, is for recommendations based on (assumed) user preferences \cite{vogiatzis2008framework, lu_leveraging_2017}. The robot in \cite{quindere2013evaluation} is able to acquire and store knowledge, which is explicitly declared. In some cases the interaction with the robot is used to teach it new actions \cite{she2014back}, grasping methods in combination with object names \cite{peltason2012structuring}, or following instructive tasks \cite{rybski2007interactive}. Other forms of knowledge acquisition include active learning from past interactions \cite{sugiura2010active}, the internet \cite{nuccio_interaction_2018} or updates of the world model \cite{chen2010developing} in order to avoid failures of the interaction. Another case of knowledge acquisition takes place when the robot learns personal information about the human and stores it in a knowledge base \cite{edirisinghe2018application,alonso-martin_multimodal_2013}.

\subsection{Evaluation methods of DM in HRI}
\label{subsec:Evaluation}
Different techniques have been used for evaluating the performance of dialogue managers in general. A detailed survey of them can be found in \cite{deriu2021survey}. For the evaluation of dialogue managers in HRI, subjective and objective measures can be used (see table~\ref{tab:eval}). Performing user studies for the evaluation has the advantage that they allow to evaluate the performance of the dialogue manager and robot in a real interaction, while allowing the users to give their opinions in questionnaires, like the Godspeed questionnaire \cite{bartneck2009measurement}. That way, user frustration and other subjective measurements can be included in the evaluation. However, user studies are resource- and time-intensive. In the reviewed papers, 35 user studies were reported, with a few (3) of those being in-the-wild studies \cite{campos2018challenges, bohus2014managing,johansson2015opportunities}. Other evaluation methods include simulations (7), evaluation based on a corpus of previously gathered data (10), while some papers just describe possible  interaction scenarios (17). The participant numbers reported for the user studies differ greatly, ranging from 2 to 97 participants, with an average of 21 participants (SD 18.4). Three studies \cite{gorostiza2010natural,sugiura2010active,lu_leveraging_2017} mention tests with users, but do not report any specifics about the number of participants. For in-the-wild studies it is more difficult to report exact participant numbers, since it is not always clear if every user interacted only once, which is why \cite{campos2018challenges, bohus2014managing,johansson2015opportunities} report the total number of recorded interactions instead.

For the evaluation of DM in HRI, task success is a common metric, since the dialogue and task are often tightly coupled in task-based systems and the dialogue is needed to achieve the goal. Since 58 of the 68 papers report on task-based interactions, task success offers an easily accessible option to evaluate the interaction. Due to the integration of the dialogue manager into a bigger system, the dialogue manager is often evaluated in combination with the other modules of the robot. For example, the task success can include the dialogue, but also the motor functions, to achieve the task.
While the integration of dialogue managers into a robot opens up the possibility of using evaluation methods that asses the whole system, it complicates the evaluation. An example is that the dialogue manager gets the information from the speech recognition module, which can introduce misunderstandings that the dialogue manager then has to deal with. To avoid those problems, simulations or corpus-based evaluations can be used. An evaluation of the dialogue manager alone does not necessarily reflect how it would work in combination with the robot's other parts, especially if the robot makes use of multimodal data.

Number of turns is a metric that can be used system-independent and would aid in comparability of different systems and tasks if reported for all studies. For task-based studies we observed a high number of papers reporting the task-success rate; a metric also observed as a common evaluation method by \cite{deriu2021survey}. However, the evaluation of the system is still depending on the type of system used \cite{deriu2021survey}. Even though, this can partly account for the high number of different evaluation metrics, it would be advantageous if common metrics, like the number of turns, used would be integrated into the evaluation of all studies, to make them more comparable.

\begin{table}[t]
    \centering
    \caption{Subjective and objective measures used for evaluating DM in HRI.}
    \begin{tabular}{l}
    \toprule
    Subjective measures\\
    \midrule
    User satisfaction \cite{campos2018challenges,thomason2015learning,devillers_multimodal_2015} \\
    User frustration \cite{amiri2019augmenting,thomason2015learning} \\
    Perceived interaction pace \cite{cuayahuitl2014nonstrict} \\
    Perceived understanding \cite{lison2015hybrid,thomason2015learning} \\
    Scales from social psychology/communication \cite{torrey2006effects} \\
    Naturalness and pleasantness \cite{lison2015hybrid, lucignano_dialogue_2013} \\
    Ease of use \cite{cuayahuitl2014nonstrict, thomason2015learning}\\
    Likability of the robot \cite{gieselmann2007problem,torrey2006effects,dino2019delivering,lala2017attentive,li2006dialog}\\
    (Modified version of) the Godspeed questionnaire \cite{milhorat2019conversational, foster2012two} \\ 
    Items from the Interpersonal Communication Satisfactory Inventory \cite{campos2018challenges} \\
    Perceived common ground \cite{chai2014collaborative}\\
    Likelihood of usage in the future \cite{amiri2019augmenting, cuayahuitl2014nonstrict, thomason2015learning}\\
    Appropriateness of the answer \cite{lison2015hybrid,gervits2021classification}\\
    \\
    
    \toprule
    Objective measures\\
    \midrule
    Number of turns \cite{chai2014collaborative,lison2015hybrid,thomason2015learning,zhang2015corpp,gieselmann2007problem,lison2013model,foster2012two,lee2010hybrid, holzapfel_dialogue_2008, prommer_rapid_2006}\\
    Number of repair turns \cite{lison2015hybrid, li_computational_2006}\\
    Number of fallback utterances \cite{milhorat2019conversational}\\
    Number of rejected utterances \cite{quindere2013evaluation}\\
    Precision and recall
    \cite{lala2018evaluation,milhorat2019conversational}\\
    Accuracy \cite{amiri2019augmenting,lison2012probabilistic,hemachandra2015information,zhang2015corpp,shen2016speech,lala2018evaluation}\\
    F1 scores \cite{amiri2019augmenting,lala2018evaluation} \\
    Task success rate \cite{chai2014collaborative,amiri2019augmenting,lee2010hybrid,quindere2013evaluation,thomason2015learning,gieselmann2007problem,skantze2013exploring,foster2012two, holzapfel_dialogue_2008, lucignano_dialogue_2013,chao_timed_2016, prommer_rapid_2006,foster_two_2012}\\
    Entropy reduction \cite{hemachandra2014information,hemachandra2015information}\\
    Latency and false cut-in rate \cite{lala2018evaluation}\\
    Reward/Cost functions \cite{amiri2019augmenting,zhang2015corpp,doshi2007efficient,lu_leveraging_2017,prommer_rapid_2006}\\
    Dialogue duration \cite{lison2015hybrid}\\
    \end{tabular}
    \label{tab:eval}
\end{table}

\subsection{Challenges}
\label{subsec:Problems}
Dialogue managers which are integrated into a robot do not act independently, but are part of a bigger architecture. While this is also true for dialogue managers used without robots, robots often make use of additional sensors and actuators to integrate multimodal in- and output into the system. Currently, there is no commonly used off-the-shelf solution for spoken human-robot interaction, but instead there is a variety of different frameworks, not only for dialogue management, but also for the related modules. This variety makes it challenging to select the most appropriate solution in cases where an existing framework is used.
Due to the tight coupling of modules in a robot, the dialogue managers are also influenced by the problems of other modules, such as speech recognition errors
\cite{inoue2020attentive,lala2017attentive,funakoshi2007robust,quindere2013evaluation}. Apart from improving the speech recognition module directly, it is possible to include mechanisms into the dialogue manager that are responsible for dealing with speech recognition errors \cite{funakoshi2007robust}.

Talking to a robot raises expectations about their conversational capabilities. When those expectations are not met, the user can become frustrated \cite{amiri2019augmenting}.
Expectations are shaped in part by the morphology of the robot \cite{kunold2023not}, making the consideration of the morphology on the expected conversational capabilities necessary. For example, \cite{milhorat2019conversational} use the android robot ERICA and state that ``Erica’s realistic physical appearance implies that her spoken dialogue system must have the ability to hold a conversation in a similarly human-like manner by displaying conversational aspects such as backchannels, turn-taking and fillers''. The robot morphology does not only affect the expectations, for example regarding the human-likeness of its conversational capabilities, but also limits the tasks the robot can do, impacting the domains the robot has to be able to converse about.

In human-human conversations, common ground plays an important role and helps to reduce miscommunication \cite{stalnaker2002common}. In spoken HRI this is even more challenging since the perception of humans and robots is not the same. Expectations based on human-human conversations do not necessarily hold for human-robot conversations. This can lead to a gap between the perceived versus the real common ground \cite{chai2014collaborative}.

Interactions with robots are not purely speech based, but they can also make use of multimodal cues like gestures, gaze or facial expressions. When the robot is using multimodal input for deciding the next dialogue move, it is not enough to simply detect the multimodal cues, but they also need to be integrated for further processing. A separate model for multimodal fusion can be used that then forwards the fused information to the dialogue manager, as is done for example in \cite{holzapfel_dialogue_2008, lucignano_dialogue_2013,alonso-martin_multimodal_2013, li_computational_2006}. Before integrating multimodal cues, a decision has to be made regarding the required modalities to decide which ones are necessary or expected in the specific type of interaction and should be used by the robot. The robot morphology is impacting the multimodal cue generation as well, as robotic heads, for example, cannot generate pointing gestures. Therefore, the robot's appearance is adding both restrictions and expectations and should be chosen carefully.

When a robot is placed in an environment, it can encounter the same person multiple times, or have longer interactions with the people, for example when acting as a guide \cite{vogiatzis2008framework,lu_leveraging_2017}.
In longer conversations or long-term interactions with multiple conversations, repetitions over time can annoy the user \cite{rosenthal2010mixed}. To solve this problem in long-term interactions, \cite{rosenthal2010mixed} suggest different options. For example, producing a number of different utterances for the same context which could reoccur, and remembering previous utterances or even whole conversations. To understand the robot's behaviour and its decisions, additional explanations might be required \cite{she2014back,dino2019delivering}. Especially if the robot fails without an explanation the user might misinterpret the situation or cause the failure again, due to the lack of explanation. 

However, this leads to the next problem, which is the resources needed for designing complex dialogue managers. For example, for handcrafted approaches the structure has to be defined a priori \cite{gorostiza2010natural, jiang2011configurable}, knowledge has to be integrated \cite{ion2020dialog} and even then it will not be able to cover unknown situations \cite{quindere2013evaluation,hemachandra2014information}. 
In contrast to handcrafted approaches, data-driven ones need less human effort, but rely on data for learning conversational behaviour. However, data collection in HRI is expensive \cite{doshi2007efficient} and for supervised learning expert feedback is needed \cite{lison2012probabilistic}.
It is possible that the human and the robot follow different conversational strategies, for example, one tries to exploit previous conversation topics, while the other one tries to explore new topics. If the robot does not notice those differences in strategies and rigidly follows its own strategy, it can lead to high tension in the dialogue \cite{campos2018challenges}.
Collecting data in an HRI setting is a challenging task, since the data is often multimodal and the interaction has to take place in a situated setting. This means that the data collection requires time, human participants and a system that can record all the required signals. However, if a specific robot is used for recording data, the generalizability to other robots is still unclear.

The reporting of user studies and their results is still lacking uniformity. An example for this are the multiple ways of which length can be assessed in the context of dialogues. Some papers report the number of turns \cite{holzapfel_dialogue_2008,campos2018challenges,chai2014collaborative}, while others focus on the length of turns \cite{lee2010hybrid,dino2019delivering} or the total amount of time \cite{johansson2015opportunities}. While the heterogeneous reporting on the length can make comparisons challenging, there are also still papers that do not report on the interaction length at all. 

Human-robot interaction offers the additional challenge of grounding the interaction in the environment the robot is physically located in. This is especially the case for multimodal interaction where the referral of objects can happen, for example, through gestures \cite{aviles2010integrating,aguilar_integrating_2009} or gaze \cite{skantze2013exploring}.  A robot that is referring to locations around itself, needs to know its own position as well as the names and places of the locations around it \cite{nakano2010grounding,funakoshi2007robust,hemachandra2014information,han_herme_2013}. The physical environment of the robot also contains objects, which the robot or human can refer to. Especially for robots that can perform object manipulation, it is important to be able to understand and produce references to those objects \cite{peltason2012structuring, sugiura2010active, she2014back,li_computational_2006,lison2013model}. Grounding interactions in the environment is especially challenging since the environment can change over time, with objects being placed and removed, people appearing and disappearing, and also the robot itself moving in the environment.

To date, handcrafted dialogue management approaches are still very common in HRI, due to lack of data, the aforementioned challenges due to the physical embodiment, but also because of the advantages it offers.
If the robot's conversation domain is small and transparency is required, a handcrafted approach might be a good option, whereas a probabilistic approach has the advantage that the actions can be learned from real interaction data, leading to less expert knowledge being required.

\section{Discussion}
\label{chap:discussion}
Our survey of literature that details DM in the context of HRI research revealed a variety of used DM approaches, influenced by the task the robot should perform and the needed reliability and capabilities. In this section, we will discuss the key challenges that have to be faced to make progress in dialogue management for social robots.

For task-based human-robot dialogues, the majority of systems are using handcrafted approaches (see section~\ref{subsec:Capabilities}) among which model-based approaches are the most popular (see fig.~\ref{fig:DM}). End-to-end approaches, like they are used for non-HRI task-based DM \cite{zhao2019review}, have not been seen at all in the included papers. The resources needed for designing complicated dialogue managers constitute a challenge. For handcrafted approaches the structure has to be defined \textit{a priori} \cite{gorostiza2010natural,jiang2011configurable}, knowledge has to be integrated \cite{ion2020dialog} and even then they will not be able to cover unknown situations \cite{quindere2013evaluation,hemachandra2014information}. In contrast to handcrafted approaches, data-driven dialogue managers need less human effort, but rely on data for learning conversational behaviour. However, data collection in HRI is expensive \cite{doshi2007efficient} and  expert feedback is needed for supervised learning \cite{lison2012probabilistic}. 

In human-human conversations, common ground plays an important role and helps to reduce miscommunication \cite{stalnaker2002common}. In spoken HRI this is even more challenging since the perception of humans and robots is not the same. Expectations learned from human-human conversations do not necessarily hold for human-robot conversations. This can lead to a gap between the perceived versus the real common-ground \cite{chai2014collaborative}. The physical presence of robots has an effect on the interaction \cite{li2015benefit}, however, it is not clear how exactly those influences look like for the different combinations as they are illustrated in \ref{fig:rotating}. This makes it challenging to transfer knowledge gained from human-human or even human-agent interaction to human-robot interaction.

While the possible interaction domains for DM in HRI span a wide area and can be combined with varying robots and environments (see fig.~\ref{fig:rotating}), most of them are task-based, for example for deliveries or navigation. Being in control of the interaction seems to be an important factor for DM in HRI, which makes handcrafted approaches a compelling option, especially in task-based interaction domains. One trend we observed is that hybrid approaches divide the dialogue manager into smaller expert modules that are responsible for specific tasks within the dialogue manager. The different experts use different DM strategies and approaches themselves, based on the task that is assigned to them. This overall strategy of distributing the task to experts leads to more flexibility in the construction of the interaction, since the parts that are responsible for different tasks can use the approach that is best for that specific task.

A capability that is specific to DM in robots, is the grounding of the dialogue in the physical environment. For DM in HRI, the user and the robot are both physically situated in the environment, which can take different forms (see fig.~\ref{fig:rotating}). Because of this, they are exposed to changes, for example, due to people moving or objects being moved. A robot can encounter new or already known people \cite{edirisinghe2018application} or unknown objects \cite{peltason2012structuring} during the conversation. This also relates to the question about endowing robots with a memory. More specifically, the physical situatedness of robots raises the question of how to manage such environmental dynamics in a robot’s memory. Due to their situatedness in the real world, interactions can happen more than once or for a longer time interval with the same robot. To solve the problem of repetitions over time in long-term interactions, \cite{rosenthal2010mixed} suggest different options to avoid repetition: producing a number of different utterances for the same context which could reoccur, either in a rule-based way or by deploying machine learning-based natural language generation, and remembering previous utterances or even whole conversations. The question then becomes how to endow a robot with an effective memory to sustain such long-term interactions.

Due to its physical presence in the environment, the robot's appearance has to be taken into account as well (see fig.~\ref{fig:rotating}). When talking to a robot, their physical appearance raises expectations about conversational capabilities, while not making explicit which ones are actually present, and users can easily become frustrated when those expectations are not met. Therefore, to understand the robot’s behaviour and its decisions, additional explanations will be required \cite{dino2019delivering,she2014back}. To avoid future failures, the robot should be transparent about the reason of a failure when it happens, so that the user can try to avoid it in the future.
The implications the chosen robot morphology has on the dialogue, is rarely discussed in the surveyed papers. Without this information, it is difficult to judge if and how the obtained results are transferable to a different robot. If the choice of robot and the resulting effects on the interaction would be more commonly discussed in research papers, this would help other researchers in choosing an appropriate robot appearance for their interactions.

Dialogue managers that are integrated into a robot are part of a bigger system. Social robots are typically multimodal and therefore depend on modules, such as sensors, while influencing other modules. Which modalities are included varies from case to case, as indicated in fig.~\ref{fig:rotating}. By influencing the actuators of the robot, the dialogue can have an immediate impact on the environment. The dialogue manager does not only interact with the NLU and NLG modules, but is often also linked to other modalities that, for example, manage the perception and production of multi-modal cues, like gestures or gaze. The dialogue manager is in addition influenced by the problems of certain modules, such as speech recognition errors \cite{quindere2013evaluation,funakoshi2007robust,lala2017attentive,inoue2020attentive}. Apart from improving the speech recognition module directly, it is possible to include mechanisms into the dialogue manager that are responsible for dealing with speech recognition errors \cite{funakoshi2007robust}. Due to speech recognition difficulties, multiple common capabilities focus on types of repair (see section~\ref{subsec:Capabilities}). However, while repairing problems in speech recognition with dialogue management techniques is an option, making the reasons for the failures explicit can help to improve the speech recognition on an HRI context. Speech recognition in HRI is impacted by noise of the robot's motors and the environment the robot is placed in, especially during experiments outside of the lab. Robots can encounter multiple people, both at the same time and at different times, whose age, accent and way of speaking can vary. All of those factors influence the performance of the speech recognition module.

During conversations, it is problematic if the robot loses track of the user or is ignoring multimodal cues \cite{li2006dialog}, while the human expects the robot to be able to process them. Since the dialogue manager is integrated into the robot’s architecture, the whole system should be taken into account for the evaluation. This means that it is not enough to evaluate the dialogue manager in a decoupled manner. Rather, it is necessary to consider the effects of the integration into the robot.  Using real users for evaluations comes with the advantage that it provides a clearer picture of how the system performs in actual interactions.

Several of the challenges we outlined relate to the variance of interaction domains, appearances, modalities and environments in HRI that need to be incorporated into the DM approach. One of these challenges is the integration into a bigger architecture, since problems in the other modules that relate to the multi-modal features of several robot types can impact the dialogue management and need to be accounted for. Another challenge is the lack of datasets for human-robot interaction, to further evolve probabilistic and hybrid approaches to DM in HRI. The data requirements are not only related to the domain of the interaction, but also to the type of robot and the location of the interaction, since those can impact the conversation. The problem of lack of data could be addressed by using large language models, that only have to be fine-tuned or can be directly prompted including recent turns in the conversation. However, they still have difficulties with situational awareness and the integration of the robot's sensors and actuators \cite{billing2023language,zhang2023large}. Human-robot datasets cannot easily be substituted by human-human datasets, since robots have different abilities than humans and lack most of the common sense abilities humans have. Datasets, moreover, even need to take specific robot shapes and forms into account, as the capabilities present in different robot platforms differ from each other. It is therefore important to maximise transparency about the robot’s conversational capabilities.

HRI researchers new to DM can make an informed decision regarding the selection of a dialogue management approach by assessing which of the presented approaches fits their requirements and limitations best. Based on our observations, hybrid approaches are a good option for more complex dialogue managers where the dialogue is an essential part of the interaction. If the interaction is simpler from a dialogue perspective, or not enough data is available, handcrafted approaches, especially model-based ones, are a viable option. While a variety of robots with a range of different shapes and appearances has been used for spoken HRI so far, it would be helpful if the implications of robot platform choices on the dialogue would be discussed more extensively. Especially due to the situatedness in the environment, the required multimodal inputs need to be taken into account in advance. As a start we would suggest to include only those, however, that are necessary for the interaction to not overly complicate the dialogue management system. Using already used evaluation metrics, especially those that are easy to record, would help to compare performance of the systems used.

Even though dialogue is common in human-robot interactions, it is rarely the main focus of the interaction design. In current research, dialogue is often only seen as a tool to achieve a task with a robot. Moreover, in papers that do take a dialogue management perspective, the dialogue manager is typically evaluated in non-embodied agents, which neglects robot-specific challenges that need to be addressed. In order to more structurally address these challenges, it is important that the best of both fields is combined to develop a more standardised approach for DM that can be drawn upon in the diverse interaction scenarios that arise in HRI studies. 

\begin{acks}
This research was (partially) funded by the Hybrid Intelligence Center, a 10-year programme funded by the Dutch Ministry of Education, Culture and Science through the Netherlands Organisation for Scientific Research, \url{https://hybrid-intelligence-centre.nl}, grant number 024.004.022.
\end{acks}

\bibliographystyle{ACM-Reference-Format}
\bibliography{DMinHRI_ref}

\end{document}